\documentclass[letterpaper, 10 pt, conference]{ieeeconf}  

\IEEEoverridecommandlockouts                              

\usepackage{graphicx}
\usepackage{amsmath}
\usepackage{amssymb}
\usepackage{booktabs}
\usepackage{mathtools}
\usepackage{multirow}
\usepackage{caption}
\usepackage{subcaption}
\usepackage{algorithm}
\usepackage{algorithmic}
\usepackage{makecell}
\usepackage{xcolor}

\usepackage[capitalize]{cleveref}
\crefname{section}{Sec.}{Secs.}
\Crefname{section}{Section}{Sections}
\Crefname{table}{Table}{Tables}
\crefname{table}{Tab.}{Tabs.}

\title{\LARGE \bf
Evaluating the Adversarial Robustness of Detection Transformers
}

\author{Amirhossein Nazeri$^{*}$, Chunheng Zhao$^{*}$ and Pierluigi Pisu
\thanks{$^{*}$Equal Contribution}
\thanks{The authors are with the Department of Automotive Engineering, Clemson University, 4 Research Drive, Greenville, SC 29607, USA (e-mail: anazeri@clemson.edu; chunhez@clemson.edu; pisup@clemson.edu)}
\thanks{The manuscript is under review, all rights are preserved.}
}

\begin{document}

\maketitle
\thispagestyle{empty}
\pagestyle{empty}

\begin{abstract}

Robust object detection is critical for autonomous driving and mobile robotics, where accurate detection of vehicles, pedestrians, and obstacles is essential for ensuring safety. Despite the advancements in object detection transformers (DETRs), their robustness against adversarial attacks remains underexplored. This paper presents a comprehensive evaluation of DETR model and its variants under both white-box and black-box adversarial attacks, using the MS-COCO and KITTI datasets to cover general and autonomous driving scenarios. We extend prominent white-box attack methods (FGSM, PGD, and C\&W) to assess DETR’s vulnerability, demonstrating that DETR models are significantly susceptible to adversarial attacks, similar to traditional CNN-based detectors. Our extensive transferability analysis reveals high intra-network transferability among DETR variants, but limited cross-network transferability to CNN-based models. Additionally, we propose a novel untargeted attack designed specifically for DETR, exploiting its intermediate loss functions to induce misclassification with minimal perturbations. Visualizations of self-attention feature maps provide insights into how adversarial attacks affect the internal representations of DETR models. These findings reveal critical vulnerabilities in detection transformers under standard adversarial attacks, emphasizing the need for future research to enhance the robustness of transformer-based object detectors in safety-critical applications.

\end{abstract}



\section{Introduction}
\label{sec:intro}
Object detection, a fundamental task in computer vision, plays a critical role in the rapidly evolving fields of autonomous driving and mobile robotics \cite{balasubramaniam2022object}. This technology forms the backbone of perception systems that enable self-driving vehicles and robots to detect and locate other vehicles, pedestrians, traffic signs, and potential obstacles in real-time. The dual challenge of predicting both bounding boxes and corresponding classes for objects of interest within an image is particularly crucial in these domains, where the reliability and accuracy of such detection can have profound implications for safety and operational efficiency \cite{michaelis2019benchmarking}. 

Inspired by accomplishments of transformers in Natural Language Processing (NLP) \cite{vaswani2017attention}, the application of transformers in computer vision has gained substantial traction. Vision Transformer (ViT) \cite{ViT} and its variants have shown remarkable results in image classification, outperforming traditional Convolutional Neural Networks (CNNs). Their self-attention mechanism effectively captures complex, long-range dependencies within an image. Extending this approach to object detection, Carion et al. \cite{carion2020Detr} introduced Detection Transformers (DETR), which eliminate the need for hand-crafted components such as anchor generation and non-maximum suppression. By leveraging a simple transformer encoder-decoder structure connected to a CNN backbone for feature extraction, DETR allows for end-to-end training of object detectors with reduced computational complexity and fewer trainable parameters \cite{carion2020Detr}.

Despite deep learning's widespread success in computer vision, studies highlight the vulnerability of Deep Neural Networks (DNNs) to well-crafted adversarial inputs, which are subtly modified data that remain nearly imperceptible but cause misclassifications by DNNs. This vulnerability raises significant concerns about the reliability of DNNs in critical applications like autonomous vehicles, where misdetection of obstacles could potentially lead to accidents. Szegedy et al. \cite{szegedy2013AdvAtt} demonstrated this vulnerability by adding small crafted perturbations that deceive state-of-the-art (SOTA) DNNs into causing incorrect image classifications. Numerous adversarial attacks have since been developed, and several are reported highly effective in compromising object detection performance \cite{zhang2020learning, goodfellow2014explaining, papernotAdvAtt}. Recent studies \cite{mahmood2021robustness, shao2021adversarial} have explored the vulnerability of ViT under adversarial attacks. However, the adversarial robustness of Detection Transformer (DETR) remains underexplored, with no comprehensive evaluation of their performance under standard white-box and black-box attacks, nor any comparison between results on general object detection datasets and application-specific datasets. This gap in the literature is particularly concerning given the increasing adoption of transformer-based object detectors in critical applications such as autonomous driving and robotics. 

In this paper, we present the first comprehensive study on the adversarial robustness of DETR and its variants using three prominent adversarial attacks (FGSM, PGD, and C\&W) over two SOTA object detection datasets: MS-COCO and KITTI. While PGD and C\&W were originally designed for image classification, they have demonstrated SOTA performance on object detection models \cite{du2022detects}, with PGD even outperforming SOTA methods \cite{wang2020adversarial}. By focusing on DETR as the foundational transformer-based object detection model, we isolate and examine the vulnerabilities of its core encoder-decoder components. This approach allows us to avoid the complexities introduced by variants like Anchor-DETR \cite{wang2022anchor} and Efficient-DETR \cite{yao2021efficient}, which primarily address training convergence and efficiency rather than adversarial robustness. In summary, we present our key findings as follows: (1) Our results uncover significant vulnerabilities in DETR even to basic attacks, aligning with observations from CNN-based object detection models and vision transformers. (2) Extensive transferability analysis shows good intra-network transferability but limited cross-network transferability in black-box settings; (3) Adversarial attacks can significantly alter self-attention feature maps, indicting that the attention mechanism is not as robust to adversarial examples as we might expect.(4) Through simple yet non-trivial modifications to the loss function, we propose our novel attack designed for DETR, which achieves SOTA performance with less visible perturbations on the COCO dataset. The code will be made available upon publication.


\section{Related Work}

Since 2021, the robustness of vision transformers for image classification task has garnered significant attention, with numerous studies investigating their performance under various adversarial attacks \cite{AttackViT001}, \cite{AttackViT002}, and \cite{AttackViT003}. The first work in evaluation of transformer-based image classification robustness by Mahmood et al, \cite{mahmood2021robustness} multiple classifiers including ViTs \cite{ViT} and BiTs \cite{kolesnikov2020big} were tested  against a variety of white-box and black-box attacks such as FGSM \cite{goodfellow2014explaining}, PGD \cite{PGD}, BPDA \cite{athalye2018obfuscated}, and C\&W \cite{carlini2017towards}.  They also explored the transferability of white-box adversarial examples between traditional CNN-based and transformer-based image classifiers. While the adversarial robustness of image classification transformers has been extensively studied in the literature, the robustness of object detection transformers remains less-explored due to their intricate architectures and task's complexity.
 
Recently, Lovisotto et al, \cite{DETR_patch} demonstrated that global reasoning process in dot-product attention can be significantly vulnerable when subjected to well-crafted adversarial patch causing huge performance degradation of DETR object detector. However, a limited conclusion can be drawn on the adversarial robustness of DETR as the adversarial transferability and validation on DETR variants are missing. Moreover, digital patch attacks are of less interest due to their higher visibility, easier detectability, less transferability, and greater dependency on the patch location and input visibility that make them less practical for real-world scenarios. In 2023, Leng et al, \cite{DETR_blackbox} proposed an Object-Aware mechanism based on black-box FGSM attack on traditional object detectors and one detection transformer. However, the work did not provide a comprehensive study on the robustness of detection transformers, and lacked generalization on other adversarial attacks and comparing the results on a variety of datasets. Zhang et al. \cite{DETR_blackbox_2} evaluated robustness of various object detectors (e.g. including Faster-RCNN \cite{ren2015faster}, SSD \cite{liu2016ssd}, and Deformable DETR \cite{zhu2020deformable}) against a transferable physical attack crafted based on multi-scale attention maps. The paper's outcome regarding evaluation of robustness of DETR variant is limited to one dataset, one specific black-box attack being considered where the attack is transferred to Deformable-DETR and is not directly exploited the transformer model. 

Although a limited number of studies strived to address the adversarial robustness of transformer-based object detection models, they primarily focus on black-box settings where attacks are not explicitly designed to compromise DETR models or use patch attacks, which differ significantly from pixel-level attacks and belong to a separate category. To the best of our knowledge, our work presents the first comprehensive study that evaluates the adversarial robustness of DETR and its variants in both white-box and black-box settings. We conduct our analysis on a general object detection dataset (COCO \cite{ms-coco}) and a scenario-specific dataset (KITTI \cite{kitti_dataset}), exploring DETR's vulnerabilities across different applications. This comprehensive approach allows us to offer insights that are crucial for developing more robust transformer-based object detection models for real-world applications such as autonomous driving and mobile robotics.


\section{Preliminaries}
\label{detr_model}
In this paper, we utilize DETR baseline variants with two CNN backbones, ResNet50 and ResNet101, and their dilation versions. The models we use are DETR-R50, DETR-R50-DC5, DETR-R101, and DETR-R101-DC5, as described in \cite{carion2020Detr}. The DC5 variants include a dilated C5 stage in the last stage of the backbone and remove a stride from the first convolution of this stage. The final outputs of DETR transformer with respect to inputs $x$ and network parameters $\theta$ can be defined as:
\begin{equation}
\begin{split}
O(\theta,x) &= [O_p(\theta,x), O_b(\theta,x)] \\
&= [\text{softmax}(P(\theta, x), \text{sigmoid}(B(\theta, x)]
\label{logits}
\end{split}
\end{equation}
where $O(\theta,x)$ is the final outputs consisting of class probabilities $O_p(\theta,x)$ and coordinates of bounding boxes $O_b(\theta,x)$. $O_p(\theta,x)$ is computed by applying softmax functions to the logits $P(\theta,x)$ while $O_b(\theta,x)$ is computed by applying sigmoid functions to $B(\theta,x)$. $P(\theta,x)$ and $B(\theta,x)$ are outputs from Prediction feed-forward networks (FFNs), which are the linear modules on the top of transformer architecture and act as detection heads of DETR \cite{carion2020Detr}. They are defined as: 
\begin{equation}
\begin{cases}
P(\theta,x) = \text{LL}(h_s) \\
B(\theta,x) = \text{MLP}(h_s)
\label{logits}
\end{cases}
\end{equation}
where LL is a single linear layer network and MLP is a multi-layer perceptron network. $h_s$ represents the hidden states from final decoder layer, while $h_s^k$ represents the hidden states from the $k^{th}$ decoder layer.  

The labels are:
\begin{equation}
t = \{t_c, t_a\}
\label{labels}
\end{equation}
where $t_c$ denotes the ground-truth classes and $t_a$ denotes and ground-truth annotations.
\section{Adversarial Attacks on DETR}
The primary goals of generating adversarial images are to (1) minimize the perturbation added to the original images and (2) induce misclassifications (i.e., altering object labels or detecting non-existent objects), subject to input data constraints. In this section, misclassifications are induced through untargeted attacks, where any incorrect class predictions are considered as successful attacks. We explore various methods for generating adversarial images, extend them to DETR, and evaluate their transferability across different models. Additionally, we introduce our proposed attack specific to DETR.

\subsection{White-box Attacks Extension to DETR}
\label{sota_attacks}
In this section, we select three prominent attacking methods originally developed for image classifiers and extend them to generate adversarial examples for DETR. These attacks assume that the adversary has white-box access to the deep learning model, meaning they have full knowledge of the model structure and weights, enabling computation of both outputs and gradients. FGSM \cite{goodfellow2014explaining} is selected for its simplicity and representativeness as a basic, one-step adversarial attack. Despite its simplicity, it effectively demonstrates the concept of adversarial attacks. For the stronger multi-step iterative attack, we select PGD \cite{PGD}. C\&W attack \cite{carlini2017towards} is also considered as it performs extremely small perturbations while achieving high success rates. While newer attack methods exist \cite{AttackViT002,zhang2020learning}, we deliberately focus on these classic attacks to demonstrate that DETR models are susceptible even to basic and classic adversarial techniques, similar to CNN-based object detectors \cite{wang2020adversarial}. 


To extend these classification-based attacks to DETR, we focus on manipulating the model's class predictions. In this \cref{sota_attacks}, we use only ground-truth classes $t_c$ in our adaptations. While this approach doesn't explicitly account for bounding box predictions, it provides a foundation for understanding DETR's vulnerabilities to adversarial attacks.

The FGSM attack can be formulated as:
\begin{equation}
x_{adv} = x + \epsilon \cdot sign(\nabla_x(-J(\theta,x,t_c)))
\label{attack1}
\end{equation}
where $x_{adv}$ is the generated adversarial image, $\epsilon$ is the multiplier to control the perturbation size, and $J(\theta,x,t_c)$ is a cross-entropy loss function dependent on the neural network parameters $\theta$, inputs $x$ and targeted classes $t_c$. As an untargated attack, we set $J=-J$ to make the loss diverge.  

The PGD attack can be formulated as a multi-step FGSM:
\begin{equation}
x_{adv}^{i+1} = \Pi_{x+S}(x_{adv}^i + \epsilon \cdot sign(\nabla_xJ(\theta,x,t_c)))
\label{attack2}
\end{equation}
where $\Pi_{x+S}$ projects perturbations into the set $S$ to limit the perturbation size at each step $t$ and $L_{\infty}$ is used as the distance metric.

The C\&W attack can be formulated as:
\begin{equation}
\begin{cases}
\text{Minimize } {\|x_{adv}-x \|}_2^2 + c \cdot f(x_{adv})\\
f = \text{Max} (\text{Max}\{P(\theta,x_{adv})_j:j\neq t\} - P(\theta,x_{adv})_t, -\kappa)
\end{cases}
\label{attack3}
\end{equation}
Here, a Change of Variables \cite{carlini2017towards} method is introduced such that $x_{adv}$ is not directly optimized; instead, a new variable $w$ is optimized, formulated as $x_{adv} = \frac{1}{2}\cdot (tanh(w)+1)$. This transformation ensures that the pixel values of $x_{adv}$ remain in a valid range while allowing unconstrained optimization of $w$. The parameter $\kappa$ encourages the solver to find an adversarial instance $x_{adv}$ that is classified as class $t$ with high confidence. The constant $c$ balances between the two loss terms. 

\subsection{Adversarial Attacks Transferability (Black-box Attacks)}
In this subsection, we explore the transferability of adversarial attacks on DETR, an important aspect that can reveal potential vulnerabilities in practical applications (e.g., autonomous driving). Adversarial examples generated on one network have shown to be effective on other networks, a phenomenon known as adversarial attack transferability. This raises significant concerns for real-world deployments, as it suggests that DNNs may be vulnerable to adversarial attacks even when there is no direct access to the target network (i.e., black-box attacks). In black-box scenarios, attackers may train their own substitute model, generate adversarial examples against this substitute, and then transfer these examples to a victim model. This approach can be effective even with limited information about the victim model.

In this paper, we investigate both intra-network and cross-network transferability to further explore the potential vulnerabilities of DETR. Intra-network transferability refers to the effectiveness of adversarial examples when generated and evaluated using DETR and its variants. This type of transferability helps us understand how robust different versions of DETR are to attacks generated on similar architectures. Additionally, we conduct a preliminary cross-network transferability analysis by evaluating how adversarial examples generated with DETR perform when applied to a CNN-based object detector, specifically Faster R-CNN \cite{ren2015faster}. To quantify the effectiveness of these transfers, we formally define untargeted transferability as follows:
\begin{equation}
    TR_{m,n}= \frac{AP_{clean}^m-AP_{adv(n)}^m}{AP_{clean}^n-AP_{adv(n)}^n}
\end{equation}
where $TR_{m,n}$ represents the transfer rate of adversarial examples generated on model $n$ and tested on model $m$. $adv(n)$ denotes the adversarial examples generated using model $n$. $AP_{adv(n)}^m$ denotes the average precision evaluated on model $m$ using $adv(n)$, while $AP_{adv(n)}^n$ denotes the average precision evaluated on model $n$ using $adv(n)$.

\subsection{Our Attack on DETR}

\begin{algorithm}[tb]
   \caption{Our Attack on DETR}
   \label{alg:new_attack}
\begin{algorithmic}[1]
   \INPUT initial image $x$, ground-truth label $t$, steps $m$, perturbation constant $c$, initial perturbation constant $\alpha$.
   \OUTPUT adversarial image $x_{adv}$
   \STATE {\bfseries Initialize:}
   \STATE $x \gets x + \alpha \cdot \nabla_x(-J_{cls}(\theta,x,t_c))$
   \STATE $w_0 = zeros(x)$
   \FOR{$i=0$ {\bfseries to} $m-1$}
   \STATE $x_{adv} = \frac{1}{2}\cdot (tanh(w_i)+1)$
   \STATE $Loss_{dm} = L_2(x_{adv},x)$
   \STATE $Loss_{cls} = c \cdot f(x_{adv})$
   \STATE $Loss_{bb}^{\{o,k\}} = -J_{bb}^{\{o,k\}}(\theta,x_{adv},t_a)$
   \STATE $Loss_{iou}^{\{o,k\}} = -J_{iou}^{\{o,k\}}(\theta,x_{adv},t_a)$
   \STATE {Update $w_i$ with gradient decent \\
   $w_i \gets \nabla_{w_i}(Loss_{total}=Loss_{dm}+Loss_{cls}+Loss_{bb}^{\{o,k\}}+Loss_{iou}^{\{o,k\}})$}
   \IF {$Loss_{total}$ does not converge}
   \STATE {\bfseries return:} $x_{adv} = \frac{1}{2}\cdot (tanh(w_i)+1)$
   \ENDIF
   \ENDFOR
   \STATE {\bfseries return:} $x_{adv} = \frac{1}{2}\cdot (tanh(w_i)+1)$
\end{algorithmic}
\end{algorithm}

In this subsection, we introduce a novel untargeted attack specifically designed for DETR object detection models assuming the same white-box adversary as detailed in \cref{sota_attacks}. Our approach improves upon existing methods by combining an initial one-step perturbation with a multi-step process based on modified C\&W attacks, considering DETR's intermediate loss functions.

Our attack method, as outlined in \cref{alg:new_attack}, consists of two main stages: (1) initialize the attack with a one-step slight perturbation; (2) apply a multi-step C\&W process with modified loss functions to the slightly-perturbed images. By using slightly-perturbed images as inputs, our multi-step attack generates perturbations based on already perturbed images, tending to maintain similarity to these slightly-perturbed images rather than the pure clean images. This approach ensures that even if the iterative procedures fail to produce effective perturbations, the initial one-step attack can compensate and provide a baseline level of adversarial effect. The updated input images (i.e., slightly-perturbed images) are generated as follows:
\begin{equation}
\begin{cases}
x = x + \alpha \cdot \nabla_x(-J_{cls}(\theta,x,t_c))\\
J_{cls}(\theta,x,t_c) = J_{cls}^o(\theta,x,t_c) + \sum_{k=1}^N J_{cls}^k(\theta,x,t_c)\\
J_{cls}^{\{o,k\}}(\theta,x,t_c) = L_{CE}(\theta,x,t_c)
\end{cases}
\label{jl}
\end{equation}
where $J_{cls}^{\{o,k\}}(\theta,x,t)$ computes the cross-entropy loss between predicted and targeted classes; $J_{cls}^o$ respresents the classification loss from the final output layer and $J_{cls}^k$ denotes the classification loss from the $k^{th}$ decoder layer; $N$ is the total number of decoder layers. Gradients $\nabla_x$ are computed with respect to the negetive loss $-J_{cls}$, aiming to minimize the confidence score of the targeted classes.

After updating the input images, we leverage and customize C\&W attacks with redesigned loss functions specifically targeting DETR architectures. The optimization process minimizes $Loss_{total}$, aiming to achieve two main objectives: minimizing perturbation and inducing misclassification. The total loss is composed of four components:
\begin{equation}
\begin{split}
Loss_{total}=&\omega_1\cdot Loss_{dm}+\omega_2\cdot Loss_{cls}+\\
&\omega_3\cdot Loss_{bb}^{\{o,k\}}+\omega_4\cdot Loss_{iou}^{\{o,k\}}
\end{split}
\label{loss}
\end{equation}
The optimal weight values $\omega$ are determined using a grid search approach. Each component of the loss function is designed to achieve the following objectives:

\noindent\textbf{Reducing the difference between adversarial and input images:} The first objective aims to minimize perturbations by evaluating the $L_2$ similarity between generated adversarial images $x_{adv}$ and input images $x$ (i.e., slightly-perturbed images).
\begin{equation}
Loss_{dm} = L_2(x_{adv},x) = {\|x_{adv}-x \|}_2^2
\label{loss1}
\end{equation}

\noindent\textbf{Eliminating the detection of the targeted class.} The second objective focuses on inducing misclassifications. This objective is achieved through three sub-objectives:
\begin{itemize}
\item Maximizing the confidence score of untargeted classes.
\begin{equation}
Loss_{cls} = c \cdot f(x_{adv})
\label{loss2}
\end{equation}
where $f(x_{adv})$ is defined as in \cref{attack3}, and $\kappa=0$ is set as in the original C\&W attacks.
\item Minimizing the regression score of bounding box coordinates of targeted classes:
\begin{equation}
Loss_{bb}^{\{o,k\}} = - (J_{bb}^o(\theta,x_{adv},t_a) + \sum_{k=1}^N J_{bb}^k(\theta,x_{adv},t_a))
\label{loss3}
\end{equation}
where $J_{bb}^{\{o,k\}}(\theta,x_{adv},t_a)$ computes the $L_1$ regression loss from the output layer and $k^{th}$ decoder layer, correspondingly.
\item Minimizing the Intersection over Union (IoU) score of targeted classes:
\begin{equation}
Loss_{iou}^{\{o,k\}} = - (J_{iou}^o(\theta,x_{adv},t_a) + \sum_{k=1}^N J_{iou}^k(\theta,x_{adv},t_a))
\label{loss4}
\end{equation}
where $J_{iou}^{\{o,k\}}(\theta,x_{adv},t_a)$ computes the Generalized Intersection over Union (GIoU) loss from the output layer and $k^{th}$ decoder layer, correspondingly.
\end{itemize}

Overall, the proposed attack considers not only the outputs from the last decoder layer $h_s$ but also the outputs from the $k^{th} $intermediate decoder layers $h_s^k$, which is specifically designed for the transformer-based architecture of DETR. Unlike the attacks described in \cref{sota_attacks} that only consider object classes, the last two loss components ($loss_{bb}$ and $loss_{iou}$) in this attack aim to cause the detector to incorrectly predict object locations, potentially resulting in misdetection. While our primary goal is to mislabel the targeted classes, incorrect bounding box placement can further degrade the detector's performance.

\section{Experiments}
\subsection{Exprimental Setup}
\subsubsection{White-box attacks setup}
\label{white_box_setup}
\noindent\textbf{Datasets:} We employ two popular object detection benchmarks, the MS COCO 2017 \cite{ms-coco} and KITTI Vision \cite{kitti_dataset}. COCO is used as a general object detection benchmark with 80 categories, and our experiments are conducted on its validation set, which contains 5,000 images. KITTI is specifically used for autonomous driving or mobile robotics applications, containing 9 classes. For this experiment, we use a subset of it to build a validation set, which consists of 1871 images. These datasets are chosen to evaluate attacks on both general object detection tasks (COCO) and practical application scenarios (KITTI).

\noindent\textbf{Models:} We evaluate four DETR variants: DETR-R50, DETR-R50-DC5, DETR-R101, and DETR-R101-DC5. These models differ in their backbone architectures (ResNet-50 or ResNet-101) and the use of dilation technique (DC5). Further details can be found in \cite{carion2020Detr}.

\begin{table*}[t]
\caption{White-box attacks against DETR models on COCO and KITTI. COCO evaluation metrics $AP(IoU=0.50:0.95)$ and $AR(maxDets=100)$ are reported.}
\label{tab:white_box_results}
\centering
\renewcommand{\arraystretch}{1} 
\resizebox{\textwidth}{!}{  
\begin{tabular}{|l|l|c|c|c|c|c|c|c|c|c|c|}
\hline
\multicolumn{12}{|c|}{COCO} \\
\hline
 \multirow{2}{*}{Models} & \multirow{2}{*}{Metric} & \multirow{2}{*}{Clean} & \multicolumn{3}{c|}{FGSM} & \multicolumn{2}{c|}{PGD} & \multicolumn{3}{c|}{C\&W} & Ours\\
 \cline{4-12}
 & & & $\epsilon=0.03$ & $\epsilon=0.05$ & $\epsilon=0.1$ & $\epsilon=0.03$ & $\epsilon=0.1$  & $c=1$ & $c=3$ & $c=5$ & $c=0.8$ \\
\hline
\multirow{2}{*}{Detr-R50} & \scriptsize $AP$   & \scriptsize 0.420  & \scriptsize 0.327 & \scriptsize 0.293 & \scriptsize 0.188  & \scriptsize 0.095 & \scriptsize 0.070 & \scriptsize 0.135 & \scriptsize 0.105 & \scriptsize 0.087 & \scriptsize 0.084\\ 
& \scriptsize $AR$                             & \scriptsize 0.574  & \scriptsize 0.492 & \scriptsize 0.459 & \scriptsize 0.343  & \scriptsize 0.209 & \scriptsize 0.165 & \scriptsize 0.275 & \scriptsize 0.238 & \scriptsize 0.209 & \scriptsize 0.203\\
\hline
\multirow{2}{*}{Detr-R50-DC5} & \scriptsize $AP$  & \scriptsize 0.433 & \scriptsize 0.341 & \scriptsize 0.309 & \scriptsize 0.203 & \scriptsize 0.094 & \scriptsize 0.073 & \scriptsize 0.111 & \scriptsize 0.099 & \scriptsize 0.081 & \scriptsize 0.047\\
& \scriptsize $AR$                                & \scriptsize 0.594 & \scriptsize 0.512 & \scriptsize 0.484 & \scriptsize 0.360 & \scriptsize 0.209 & \scriptsize 0.170 & \scriptsize 0.243 & \scriptsize 0.231 & \scriptsize 0.199 & \scriptsize 0.139\\
\hline
\multirow{2}{*}{Detr-R101} & \scriptsize $AP$  & \scriptsize 0.435 & \scriptsize 0.336 & \scriptsize 0.300  & \scriptsize 0.195 & \scriptsize 0.086 & \scriptsize 0.060 & \scriptsize 0.137 & \scriptsize 0.112 & \scriptsize 0.088 & \scriptsize 0.063\\
& \scriptsize $AR$                              & \scriptsize 0.590 & \scriptsize 0.506 & \scriptsize 0.470  & \scriptsize 0.349 & \scriptsize 0.207 & \scriptsize 0.168 & \scriptsize 0.291 & \scriptsize 0.251 & \scriptsize 0.218 & \scriptsize 0.166\\
\hline
\multirow{2}{*}{Detr-R101-DC5} & \scriptsize $AP$ & \scriptsize 0.449  & \scriptsize 0.354 & \scriptsize 0.322  & \scriptsize 0.205 & \scriptsize 0.090 & \scriptsize 0.063 & \scriptsize 0.130 & \scriptsize 0.107 & \scriptsize 0.087 & \scriptsize 0.034 \\
& \scriptsize $AR$                                & \scriptsize 0.604 & \scriptsize 0.525  & \scriptsize 0.492  & \scriptsize 0.363 & \scriptsize 0.210 & \scriptsize 0.164 & \scriptsize 0.172 & \scriptsize 0.246 & \scriptsize 0.210 & \scriptsize 0.112 \\
\hline
\multicolumn{12}{|c|}{KITTI} \\
\hline
 \multirow{2}{*}{Models} & \multirow{2}{*}{Metric} & \multirow{2}{*}{Clean} & \multicolumn{3}{c|}{FGSM} & \multicolumn{2}{c|}{PGD} & \multicolumn{3}{c|}{C\&W} & Ours\\
 \cline{4-12}
 & & & $\epsilon=0.03$ & $\epsilon=0.05$ & $\epsilon=0.1$ & $\epsilon=0.03$ & $\epsilon=0.1$  & $c=1$ & $c=3$ & $c=5$ & $c=0.8$ \\
\hline
\multirow{2}{*}{Detr-R50} & \scriptsize $AP$        & \scriptsize{0.378}  & \scriptsize{0.324} & \scriptsize{0.162} & \scriptsize{0.086} & \scriptsize{0.044} & \scriptsize{0.034} & \scriptsize{0.056} & \scriptsize{0.052} & \scriptsize{0.049} &  \scriptsize{0.075}  \\
& \scriptsize $AR$                                  & \scriptsize{0.561}  & \scriptsize{0.528} & \scriptsize{0.319} & \scriptsize{0.192} & \scriptsize{0.129} & \scriptsize{0.112} & \scriptsize{0.146} & \scriptsize{0.137} & \scriptsize{0.129} & \scriptsize{0.194}   \\
\hline
\multirow{2}{*}{Detr-R50-DC5} & \scriptsize $AP$    & \scriptsize{0.352} & \scriptsize{0.209} & \scriptsize{0.176} & \scriptsize{0.097} & \scriptsize{0.040} & \scriptsize{0.031} & \scriptsize{0.064} & \scriptsize{0.050} & \scriptsize{0.034} & \scriptsize{0.064}    \\
& \scriptsize $AR$                                  & \scriptsize{0.544} & \scriptsize{0.398} & \scriptsize{0.351} & \scriptsize{0.226} & \scriptsize{0.122} & \scriptsize{0.092} & \scriptsize{0.167} & \scriptsize{0.130} & \scriptsize{0.110} & \scriptsize{0.179}    \\
\hline
\multirow{2}{*}{Detr-R101} &  \scriptsize $AP$      & \scriptsize{0.367} & \scriptsize{0.219} & \scriptsize{0.167} & \scriptsize{0.079} & \scriptsize{0.036} & \scriptsize{0.023} & \scriptsize{0.071} & \scriptsize{0.048} & \scriptsize{0.038} & \scriptsize{0.061}     \\
& \scriptsize $AR$                                  & \scriptsize{0.581} & \scriptsize{0.400} & \scriptsize{0.326} & \scriptsize{0.155} & \scriptsize{0.099} & \scriptsize{0.072} & \scriptsize{0.160} & \scriptsize{0.130} & \scriptsize{0.109} & \scriptsize{0.172}     \\
\hline
\multirow{2}{*}{Detr-R101-DC5} & \scriptsize $AP$   & \scriptsize{0.383} & \scriptsize{0.233} & \scriptsize{0.177} & \scriptsize{0.089} & \scriptsize{0.048} & \scriptsize{0.032} & \scriptsize{0.067} & \scriptsize{0.049} & \scriptsize{0.040} & \scriptsize{0.065}   \\
& \scriptsize $AR$                                  & \scriptsize{0.616} & \scriptsize{0.407} & \scriptsize{0.324} & \scriptsize{0.172} & \scriptsize{0.116} & \scriptsize{0.092} & \scriptsize{0.152} & \scriptsize{0.128} & \scriptsize{0.114} & \scriptsize{0.174}   \\
\hline
\end{tabular}}
\end{table*}

\noindent\textbf{FGSM:} $\epsilon$ is set to 0.03, 0.05 and 0.1.

\noindent\textbf{PGD:} $\epsilon$ is set to 0.03, and 0.1, with a total of 10 iterations. The $L_{\infty}$ bounds are set to $\pm 10/255$.

\noindent\textbf{C\&W:} $c$ is set to 1, 3 and 5, with a total of 200 iterations.

\noindent\textbf{Ours:} $\alpha$ is set to 0.3, and $c$ is set to 0.8, with 200 iterations.

The perturbation size is limited based on perceptible levels to ensure the adversarial examples remain visually similar to the original images. In addition to standard object detection metrics like Average Precision ($AP$) and Average Recall ($AR$), we use another metric for assessing adversarial robustness, the Robustness Score ($RS$). This metric evaluates a model's robustness against specific adversarial attacks, with higher $RS$ values indicating stronger resistance to the adversarial attack. $RS$ is defined as: $RS = AP_{adv}/AP_{clean}$, where $AP_{adv}$ and $AP_{clean}$ are average precision score with adversarial images and clean images, respectively.

\begin{figure}[htbp]
\centering
\includegraphics[width=3.2in]{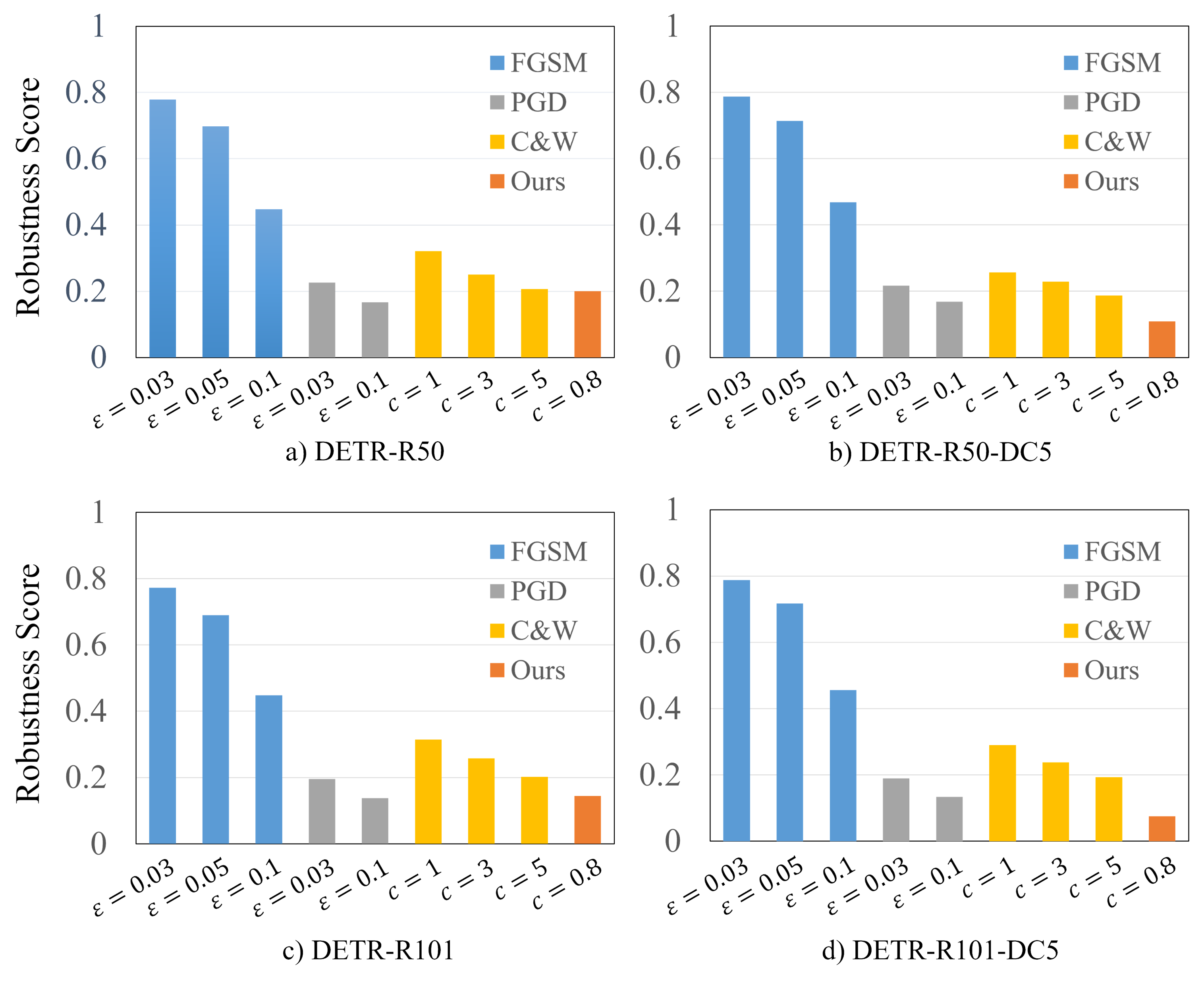}
\caption{Robustness Score of DETR variants on COCO.}
\label{Robustness_DETR_COCO}
\end{figure}

\begin{figure}[htbp]
\centering
\includegraphics[width=3.2in]{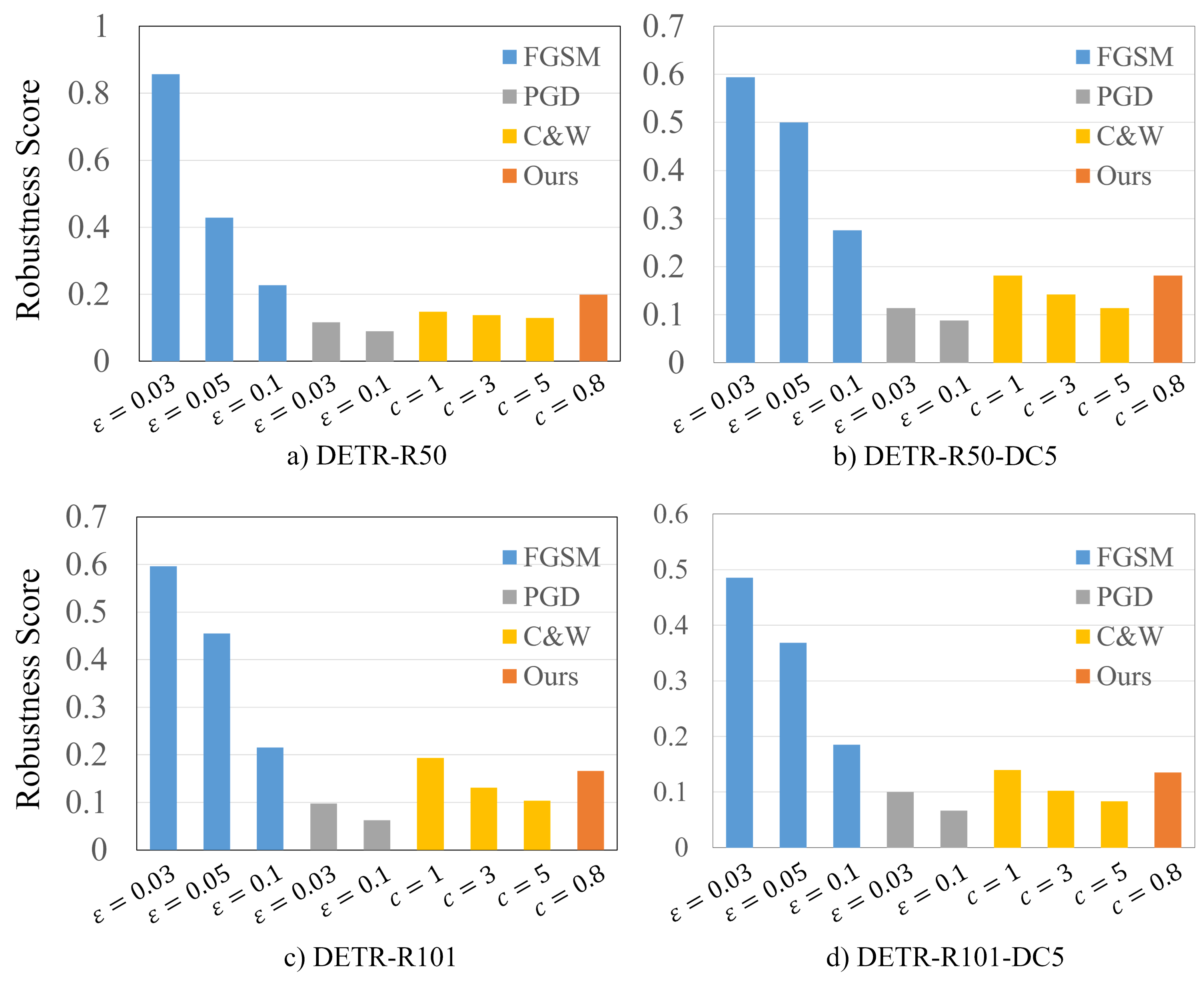}
\caption{Robustness Score of DETR variants on KITTI.}
\label{Robustness_DETR_KITTI}
\end{figure}

\subsubsection{Transferability analysis setup}
For the transferability analysis, we use the same datasets, models, and attack parameters as described in \cref{white_box_setup}. For intra-network transferability, the same four DETR variants are used. As cross-network transferability is not the main focus in this paper, only one Faster R-CNN model (Faster R-CNN-R50-FPN) \cite{ren2015faster} is used as Faster R-CNN is one of the most representative CNN-based object detectors. This allows us to evaluate transferability across distinct object detection models due to the architectural differences between transformers and traditional CNNs.

To evaluate intra-network and cross-network transferability on the KITTI dataset, we retrain the four DETR variants along with the Faster R-CNN model. Since the KITTI object detection test set is unannotated, we split the original training set into new train and validation sets using a 3:1 ratio. The four DETR variants are retrained for 25 epochs, with a transformer learning rate of $1e-5$, following the recommended recipe by \cite{carion2020Detr}. Faster R-CNN is retrained for 25 epochs, with a learning rate of $1e-3$.

\begin{table}[t]
\caption{Intra-network and cross-network transferability results on COCO and KITTI dataset. The models in different rows represent the model used to generate the adversarial examples. The models in different columns represent the model used to evaluate the adversarial examples. $TR$ is reported as the primary metric.}
\label{transfer_results}

\renewcommand{\arraystretch}{1.5}

\centering
\resizebox{\columnwidth}{!}{  
\begin{tabular}{|l|l|c|c|c|c|c|c|}
\hline
\multicolumn{7}{|c|}{\normalsize COCO} \\
\hline
\normalsize Models & \normalsize Attacks & \normalsize \rotatebox{0}{Detr-R50} & \normalsize \rotatebox{0}{\makecell[c]{Detr-R50-\\DC5}}  & \normalsize \rotatebox{0}{Detr-R101} & \normalsize \rotatebox{0}{\makecell[c]{Detr-R101-\\DC5}} & \normalsize \rotatebox{0}{\makecell[c]{Faster\\R-CNN}} \\
\hline
\multirow{3}{*}{\normalsize \rotatebox{0}{Detr-R50}} & \normalsize FGSM($\epsilon=0.05$)  &  \normalsize 100\% & \normalsize 96.8\% & \normalsize 71.8\% & \normalsize 65.3\% & \normalsize 66.9\% \\
 & \normalsize PGD($\epsilon=0.03$) & \normalsize 100\% & \normalsize 109.0\% & \normalsize 106.2\% & \normalsize 109.3\% & \normalsize 54.8\%  \\
 & \normalsize C\&W($c=3$)    & \normalsize 100\% & \normalsize 99.1\% & \normalsize 89.5\% & \normalsize 80\%  & \normalsize 66.7\% \\
\hline
\multirow{3}{*}{\normalsize \rotatebox{0}{\makecell[c]{Detr-R50-\\DC5}}} & \normalsize FGSM($\epsilon=0.05$) & \normalsize 107.5\% & \normalsize 100\% & \normalsize 78.3\% & \normalsize 70.8\% & \normalsize 70.0\% \\
 & \normalsize PGD($\epsilon=0.03$)  & \normalsize 101.5\% & \normalsize 100\% & \normalsize 101.8\% & \normalsize 104.7\% & \normalsize 51.6\% \\
 & \normalsize C\&W($c=3$)           & \normalsize 94.2\% & \normalsize 100\% & \normalsize 85.5\% & \normalsize 87.3\% & \normalsize 69.1\% \\
\hline
\multirow{3}{*}{\normalsize \rotatebox{0}{Detr-R101}} & \normalsize FGSM($\epsilon=0.05$)  & \normalsize 86.4\% & \normalsize 78.8\% & \normalsize 100\% & \normalsize 90.9\% & \normalsize 70.5\% \\
 & \normalsize PGD($\epsilon=0.03$)  & \normalsize 96.8\% & \normalsize 100.1\% & \normalsize 100\% & \normalsize 107.4\% & \normalsize 53.3\% \\
 & \normalsize C\&W($c=3$)  & \normalsize 82.5\% & \normalsize 56.1\% & \normalsize 100\% & \normalsize 86.9\% & \normalsize 59.0\% \\
\hline
\multirow{3}{*}{\normalsize \rotatebox{0}{\makecell[c]{Detr-R101-\\DC5}}} & \normalsize FGSM($\epsilon=0.05$)  & \normalsize 89.7\% & \normalsize 82.5\% & \normalsize 104.0\% & \normalsize 100\% & \normalsize 71.4\% \\
 & \normalsize PGD($\epsilon=0.03$) & \normalsize 93.6\% & \normalsize 96.1\% & \normalsize 100.3\% & \normalsize 100\% & \normalsize 51.0\% \\
 & \normalsize C\&W($c=3$) & \normalsize 79.8\% & \normalsize 81.0\% & \normalsize 91.2\% & \normalsize 100\% & \normalsize 67.8\% \\
\hline
\multicolumn{7}{|c|}{\normalsize KITTI} \\
\hline
\normalsize Models & \normalsize Attacks & \normalsize \rotatebox{0}{Detr-R50} & \normalsize \rotatebox{0}{\makecell[c]{Detr-R50-\\DC5}}  & \normalsize \rotatebox{0}{Detr-R101} & \normalsize \rotatebox{0}{\makecell[c]{Detr-R101-\\DC5}} & \normalsize \rotatebox{0}{\makecell[c]{Faster\\R-CNN}} \\
\hline
\multirow{3}{*}{\normalsize \rotatebox{0}{Detr-R50}} & \normalsize FGSM($\epsilon=0.05$)  &  \normalsize 100\% & \normalsize 77.9\% & \normalsize 73.6\% & \normalsize 59.6\% & \normalsize 84.1\% \\
 & \normalsize PGD($\epsilon=0.03$) & \normalsize 100\% & \normalsize 92.8\% & \normalsize 95.8\% & \normalsize 98.8\% & \normalsize 77.6\%  \\
 & \normalsize C\&W($c=3$)    & \normalsize 100\% & \normalsize 88.7\% & \normalsize 88.1\% & \normalsize 87.5\% & \normalsize 72.2\% \\
\hline
\multirow{3}{*}{\normalsize \rotatebox{0}{\makecell[c]{Detr-R50-\\DC5}}} & \normalsize FGSM($\epsilon=0.05$) & \normalsize 116.6\% & \normalsize 100\% & \normalsize 89.9\% & \normalsize 76.3\% & \normalsize 105.9\% \\
 & \normalsize PGD($\epsilon=0.03$)  & \normalsize 110.2\% & \normalsize 100\% & \normalsize 104.2\% & \normalsize 108.6\% & \normalsize 85.9\% \\
 & \normalsize C\&W($c=3$)    & \normalsize 107.3\% & \normalsize 100\% & \normalsize 99.0\% & \normalsize 96.0\% & \normalsize 78.8\% \\
\hline
\multirow{3}{*}{\normalsize \rotatebox{0}{ Detr-R101}} & \normalsize FGSM($\epsilon=0.05$)  & \normalsize 81.8\% & \normalsize 62.6\% & \normalsize 100\% & \normalsize 93.4\% & \normalsize 83.8\% \\
 & \normalsize PGD($\epsilon=0.03$)  & \normalsize 101.5\% & \normalsize 92.4\% & \normalsize 100\% & \normalsize 104.8\% & \normalsize 83.7\% \\
 & \normalsize C\&W($c=3$)    & \normalsize 94.7\% & \normalsize 86.3\% & \normalsize 100\% & \normalsize 100\% & \normalsize 76.0\% \\
\hline
\multirow{3}{*}{\normalsize \rotatebox{0}{\makecell[c]{Detr-R101-\\DC5}}} & \normalsize FGSM($\epsilon=0.05$)  & \normalsize 84.5\% & \normalsize 69.5\% & \normalsize 106\% & \normalsize 100\% & \normalsize 79.5\% \\
 & \normalsize PGD($\epsilon=0.03$) & \normalsize 99.7\% & \normalsize 89.8\% & \normalsize 98.5\% & \normalsize 100\% & \normalsize 82.6\% \\
 & \normalsize C\&W($c=3$)    & \normalsize 93.1\% & \normalsize 85.3\% & \normalsize 97.3\% & \normalsize 100\% & \normalsize 72.5\% \\
\hline
\end{tabular}}
\end{table}

\subsection{Quantitative Evaluation}
\subsubsection{White-box attacks analysis}
\begin{figure}[htbp]
\centering
\includegraphics[width=2.8in]{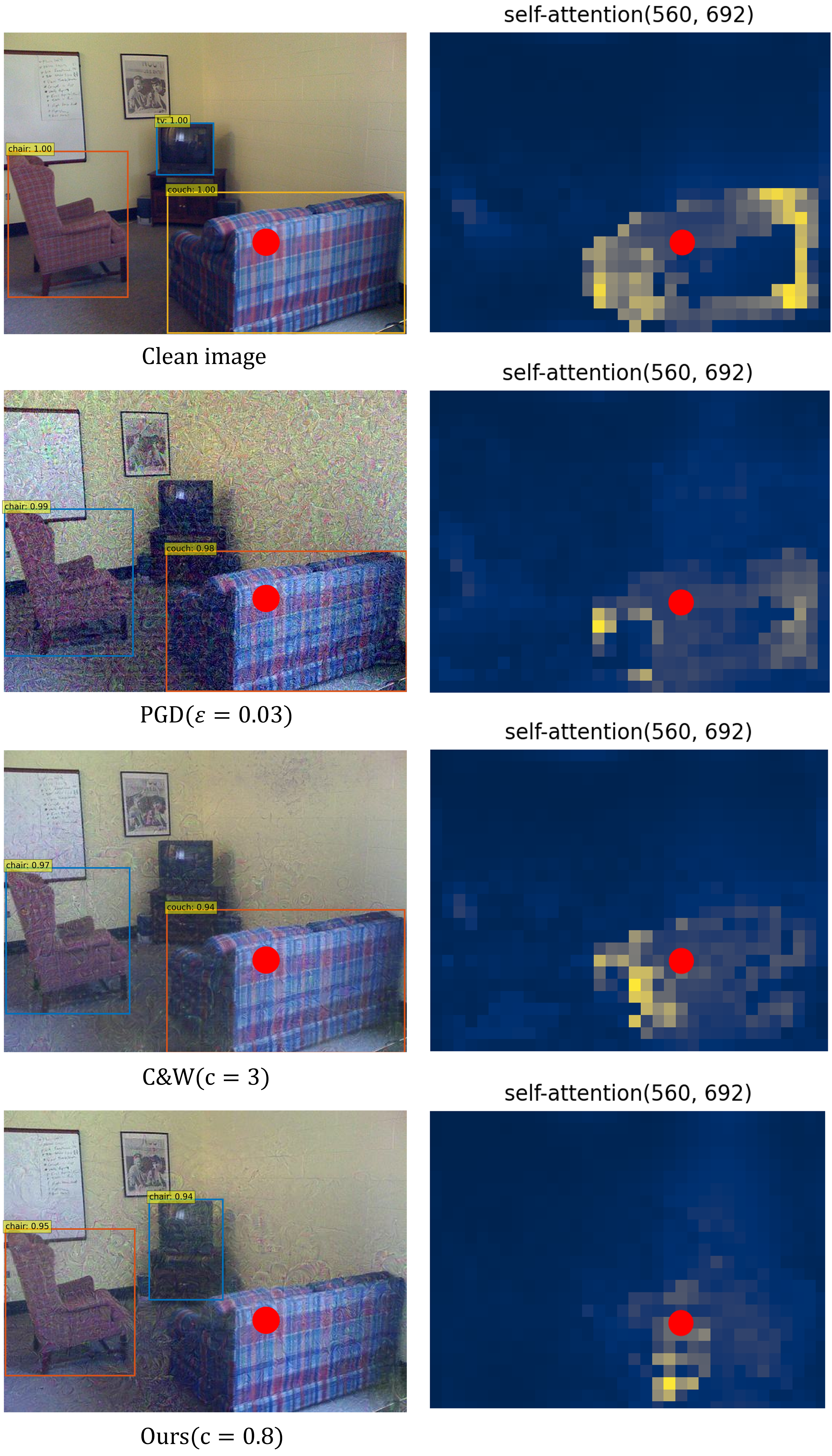}
\caption{Object detection on a sample image from COCO dataset with self-attention feature maps from the last encoder.}
\label{self_attention}
\end{figure}

\begin{figure*}[htbp]
     \centering
     \begin{subfigure}[b]{\textwidth}
         \centering
         \includegraphics[width=6in]{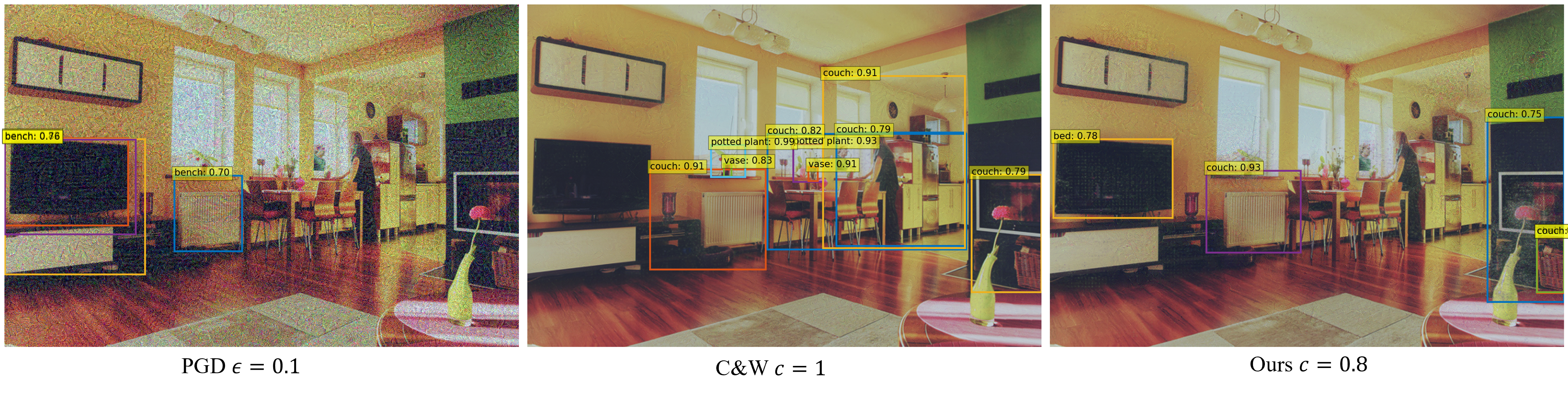}
         \caption{Attacks on COCO}
         \label{qualitative_coco}
      \end{subfigure}
      \begin{subfigure}[b]{\textwidth}
         \centering
         \includegraphics[width=6in]{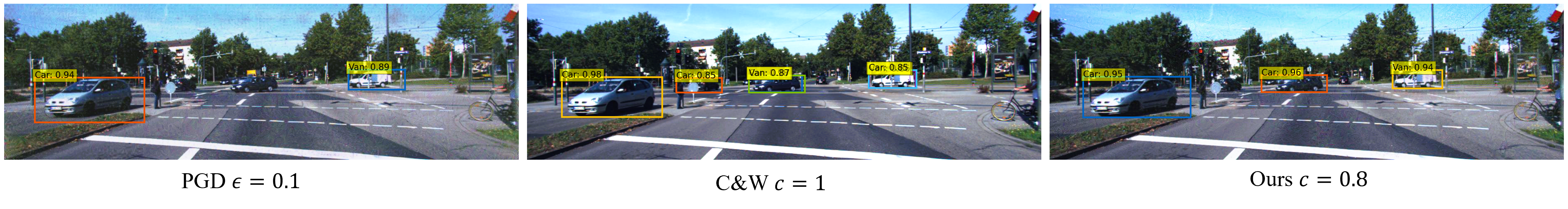}
         \caption{Attacks on KITTI}
         \label{qualitative_kitti}
      \end{subfigure}
        \caption{Object detection results on sample images under various adversarial attacks.}
        \label{qualitative_results}
\end{figure*}

We evaluate the robustness of four DETR models against four adversarial attacks on both COCO and KITTI datasets. The results are summarized in \cref{tab:white_box_results}, \cref{Robustness_DETR_COCO}, and \cref{Robustness_DETR_KITTI}, showing Average Precision ($AP$), Average Recall ($AR$) and Robustness Score ($RS$) for each model-attack combination. In summary, our results indicate that DETR models, similar to vision transformers and CNN-based object detectors, remain vulnerable to adversarial attacks. FGSM, as the most basic attacking algorithm, shows limited success rate against all four DETR models. PGD, as an iterative attack, proves to be the most effective, significantly decreasing $AP$ and $RS$. For instance, PGD attack on DETR-R50 achieves the lowest $AP$ across different attacks ($AP=0.070$ on COCO, $AP=0.034$ on KITTI). However, PGD generates relatively large perturbation sizes, as visualized in \cref{self_attention} and \cref{qualitative_results}. C\&W attack achieves a balance between performance degradation and perturbation size. These results demonstrate that DETR models have significant vulnerabilities even against classic attacks. 

In terms of our attack, it achieves SOTA performance comparable to PGD attacks with smaller perturbations on the COCO dataset. For example, our attack has $AP=0.047$ and $AP=0.034$ on DETR-R50-DC5 and DETR-R101-DC5 while PGD has $AP=0.073$ and $AP=0.060$ correspondingly. On the KITTI dataset, our attack achieves results comparable to C\&W attacks. While not as effective as PGD attacks, our method produces much smaller perturbations, as shown in \cref{self_attention} and \cref{qualitative_results}. With simple yet non-trivial modifications, our attack further explores DETR's adversarial vulnerabilities, indicating the need for developing more robust DETR models. Notably, the effects of dilation and differences between backbones are not obvious on both datasets.

\begin{table}[t]
\caption{Transferability results of our attack on COCO dataset. $TR$ is reported as the primary metric.}
\label{COCO_trasfer_ours}
\centering
\renewcommand{\arraystretch}{1}
\begin{tabular}{|l|l|c|c|c|c|}
\hline
\multirow{2}{*}{Models} & \multicolumn{4}{c|}{Detr}& Faster\\
\cline{2-5}
 & R50 &R50-DC5 &R101 &R101-DC5 & R-CNN\\
\hline
Detr-R50 & 100\%    &99.5\%	&95.8\%	& 87.1\%  &  43.5\%   \\
\hline
\end{tabular}
\end{table}

\subsubsection{Transferability analysis}
Transferability results for FGSM, PGD and C\&W attacks are presented in \cref{transfer_results}, while transferability results for our attack are shown in \cref{COCO_trasfer_ours}. In summary, attacks generated using DETR models can transfer across DETR variants but show limited transferability to Faster R-CNN models. For intra-network transferability, results demonstrate that generated adversarial examples can transfer within the DETR family, although effectiveness decreases with increasing model complexity. For example, attacks generated on DETR-R50 show decreasing performance when transferred to DETR-R50-DC5, DETR-R101, and DETR-R101-DC5. Interestingly, attacks generated on models with dilation transfer well to their base models. For instance, FGSM attacks from DETR-R50-DC5 and DETR-R101-DC5 achieve better results on DETR-R50 (107.5\% transfer rate) and DETR-R101 (104.0\% transfer rate), respectively. For cross-network transferbaility, all the attacks generated on DETR variants show limited transferability to Faster R-CNN. This suggests a potential ensemble defense technique against adversarial attacks on DETR models. The KITTI dataset shows slightly better transferability to Faster R-CNN compared to COCO, possibly due to Faster R-CNN's high clean $AP$ (0.553) on KITTI after retraining, suggesting a potential trade-off between accuracy and robustness.

Among all attacks, PGD demonstrates the best intra-network transferability, while FGSM shows the best cross-network transferability. This might be due to FGSM's one-step characteristic, which may rely less on DETR-specific gradients, allowing better transfer to non-DETR models. We argue that although FGSM is regarded as the simplest attack in the literature, it has potential benefits while conducting black-box attacks. For brevity, we conduct transferability analysis of our attack only on the COCO dataset with DETR-R50, as shown in \cref{COCO_trasfer_ours}. Our attack shows good transferability across DETR variants but limited cross-network transferability. This is expected as our attack is specifically designed for DETR, leveraging its intermediate loss functions, which differ significantly from those in Faster R-CNN.

\subsection{Qualitative Evaluation}
We provide a visual comparison of different attacks on sample images from the COCO and KITTI dataset as shown in \cref{self_attention} and \cref{qualitative_results}. Our attack achieves a small perturbation level comparable to C\&W attacks, despite employing a two-stage perturbation generation procedure. Notably, when compared to PGD attacks, our method produces significantly less visible perturbations while maintaining similar performance on the COCO dataset. In contrast to C\&W attacks, our approach provides superior performance with a similar perturbation level. Overall, our attack offers a balance between perturbation size and attack success rate.

To gain deeper insights into the impact of these attacks on DETR's internal representations, we examine the self-attention feature maps from the last encoder layer for each attack, as illustrated in Figure \ref{self_attention}. The clean image's feature map clearly highlights the entire object shape, indicating that DETR correctly focuses on relevant image regions for object detection. However, both PGD and C\&W attacks cause a noticeable shrinkage of the highlighted regions in the feature maps, suggesting successful disruption of DETR's attention mechanism. Remarkably, our proposed attack results in a feature map with the least salient features, indicating that it most significantly impacts the model's predictions. The ability of our attack to significantly alter the self-attention feature map while maintaining small perturbations underscores its effectiveness in exploiting DETR's architectural vulnerabilities. These results lead us to conclude that the attention mechanism in DETR, contrary to expectations, does not provide sufficient protection against adversarial attacks.

\section{Conclusion}
The expanding use of detection transformers in critical applications, such as autonomous driving and robotics, raises significant concerns about their security and reliability. In this paper, we conducted the first comprehensive study on the adversarial vulnerability of DETR and its variants. We extended one basic adversarial attack (FGSM) and two classic yet strong attacks (PGD and C\&W) to DETR models. Our results revealed substantial vulnerabilities in DETR models even against basic and classic attacks, consistent with the vulnerabilities observed in CNN-based object detection models and transformer-based image classification models. In addition, we found strong transferability of adversarial examples generated by DETR models across different DETR variants, indicating shared vulnerabilities within similar architectures. However, the transferability to CNN-based object detectors like Faster R-CNN is limited. This highlights that employing an ensemble of diverse models may help mitigate the impact of adversarial attacks on DETR models.

We also introduced a novel adversarial attack leveraging the intermediate loss functions of DETR. Our attack is simple yet effective, causing significant performance degradation with less visible perturbations. Our overall results, including insights from self-attention map features, exhibited that the attention mechanism cannot provide additional protection against adversarial attacks in object detection tasks, contrary to expectations. These findings have important implications for the deployment of DETR models in safety-critical applications such as autonomous driving and robotics, where robustness to adversarial attacks is crucial. Future work could focus on exploring attacks transferability to single-stage object detector (e.g., YOLO) and other DETR variants. In addition, analyzing the impact of attacks on autonomous driving systems with DETR-based perception modules would offer valuable insights for real-world applications.

\section*{Acknowledgement}
This work is funded by the National Science Foundation CNS No. 2200457. Any opinions, findings, and conclusions or recommendations expressed in this material are those of the author(s) and do not necessarily reflect the views of the National Science Foundation.

\bibliographystyle{IEEEtran}
\bibliography{root}

\end{document}